\documentclass{article}

     \usepackage[nonatbib, preprint]{neurips_2020}

\usepackage[utf8]{inputenc} 
\usepackage[T1]{fontenc}    
\usepackage{hyperref}       
\usepackage{url}            
\usepackage{booktabs}       
\usepackage{amsfonts}       
\usepackage{nicefrac}       
\usepackage{amsmath}
\usepackage{microtype}      
\usepackage{algpseudocode}
\usepackage{algorithm}
\usepackage{graphicx}
\usepackage[dvipsnames]{xcolor}

\title{Improving Layer-wise Adaptive Rate Methods using Trust Ratio Clipping}

\author{%
  Jeffrey Fong\thanks{Equal Contribution} , Siwei Chen\footnotemark[1] , Kaiqi Chen\footnotemark[1] \\
  School of Computing\\
  National University of Singapore\\
  \texttt{\{jfong, siwei-15, kaiqi\}@comp.nus.edu.sg} \\
}

\begin{document}

\maketitle

\begin{abstract}
  Training neural networks with large batch is of fundamental significance to deep learning. Large batch training remarkably reduces the amount of training time but has difficulties in maintaining accuracy. Recent works have put forward optimization methods such as LARS and LAMB to tackle this issue through adaptive layer-wise optimization using trust ratios. Though prevailing, such methods are observed to still suffer from unstable and extreme trust ratios which degrades performance. In this paper, we propose a new variant of LAMB, called LAMBC, which employs trust ratio clipping to stabilize its magnitude and prevent extreme values. We conducted experiments on image classification tasks such as ImageNet and CIFAR-10 and our empirical results demonstrate promising improvements across different batch sizes.
\end{abstract}

\section{Introduction}
The recent trend towards large scale datasets \cite{deng2009imagenet, abu2016youtube8m, sun2020waymo} requires training large neural networks to learn effectively. However, employing such large neural networks incurs the cost of larger computational power requirements and additional training time. For example, previous work took 29 hours to train ResNet-50, a state-of-the-art deep learning model, on 8 Tesla P100 GPUs \cite{he2016resnet}. Therefore, many types of optimization techniques have been proposed to accelerate training large deep neural networks. Some works have focused on data-parallel optimization where each global minibatch of data is distributed among the workers \cite{krizhevsky2014one, goyal2017accurate, li2014paramserver}, while some others have been involved in model-parallel methods \cite{shoeybi2019megatron, rajbhandari2019zero}.

One prominent type of technique involves large-batch optimization whereby gradients are computed on large minibatches in parallel. Such techniques has seen a resurgence recently due to advances in hardware capabilities, and has been shown in previous works to be able to accelerate large deep neural network training. For example, Goyal et al. \cite{goyal2017accurate} successfully trained ResNet-50 in 1 hour on 256 GPUs using distributed Stochastic Gradient Descent (SGD) with 8K minibatch size. However, such methods also underscore the need for adaptive learning rate mechanisms for large batch training. To address this need, recent work implemented layerwise adaptive learning rates for large batch training. The most successful ones are LARS \cite{you2017lars} and LAMB \cite{you2019lamb}, which calculate the trust ratio (ratio of L2-norm of weights over L2-norm of gradients) of each layer in the network. LARS and LAMB has been shown to be able to scale ResNet-50 and BERT \cite{devlin2018bert} models up to batch size of 32K without loss of accuracy, while drastically reducing the training time.

Though prevailing, such layerwise adaptive methods are observed to still suffer from unstable and extreme trust ratios which degrades performance. This happens when the weight norm becomes too large compared to the gradient norm, resulting in possible divergence. To this end, we propose an approach that entails clipping the trust ratio within a range of values. Inspired by recent work \cite{nvidia2020larc} that proposed trust ratio clipping on LARS, we propose a new variant of LAMB, called LAMBC, that clips the trust ratio for LAMB.

\paragraph{Contributions.} Our contributions in this paper are twofold: (1) we develop a new variant of LAMB, called LAMBC, for achieving stability and improvement in performance over standard LAMB, and (2) we demonstrate the effectiveness of trust ratio clipping across different image classification tasks such as ImageNet and CIFAR-10.

\section{Background}
Many neural networks can be trained using Stochastic Gradient based methods, which follows the following equation:

\begin{equation}
w_{t+1} = w_t - \eta_tu_t
\end{equation}
where $\eta_t$ is the learning rate and $u_t$ is the update at time step $t$. $u_t$ differs between different optimizers. For example, in SGD, $u_t=\frac{1}{B}\sum_{i=1}^B\nabla L(w_t)$, while in Adam \cite{kingma2014adam}, $u_t=\frac{m_t}{\sqrt{v_t}+\epsilon}$. To enable training with large batch, one way is to adjust the learning rate LR. However, the main obstacle for such a method is the instability of training with high LR. Goyal et al. \cite{goyal2017accurate} proposed to use LR warm-up which entails starting with small LR and gradually increasing LR to the target. However, such methods require manual adjustments of the LR (e.g.: rate of increase of LR and target LR in LR warm-up, etc.). Furthermore, such methods are unable to maintain the accuracy for batch size larger than 8K. Such problems lead to the layerwise adaptive methods proposed by \cite{you2017lars, you2019lamb}.

\subsection{Layerwise Adaptive Methods}
In layerwise adaptive methods, the general strategy is to perform layerwise normalization, where each layer's update is normalized to unit L2-norm. This is performed in the form $\frac{u_t^{(i)}}{||u_t^{(i)}||}$ where $i$ refers to the $i$-th layer. Similarly, the learning rate is also scaled layerwise by $\phi(||w_t||)$ for some function $\phi:\mathbb{R}^+\rightarrow\mathbb{R}^+$. Thus, the modifications result in the following weight update rule:

\begin{equation}\label{eq:layerwise_update}
w_{t+1}^{(i)} = w_t^{(i)} - \eta_t\frac{\phi(||w_t^{(i)}||)}{||g_t^{(i)}||}g_t^{(i)}
\end{equation}
where $g_t^{(i)}$ are the gradients of the $i$-th layer. For LARS, $g_t^{(i)}=m_t^{(i)}$ where $m_t^{(i)}$ is the first moment. For LAMB, $g_t^{(i)}=\frac{m_t^{(i)}}{\sqrt{v_t^{(i)}}+\epsilon}$, where $v_t^{(i)}$ and $\epsilon$ are the second moment and a small offset respectively. Eq. \ref{eq:layerwise_update} introduces a new term $\frac{\phi(||w_t^{(i)}||)}{||g_t^{(i)}||}$ which is called trust ratio. The trust ratio is essentially a ratio of the L2-norm of weights over the L2-norm of gradients. Intuitively, this offers a major benefit for large batch training. Such a normalization provides robustness to exploding and vanishing gradients since the trust ratio explicitly compares the magnitudes of the weights and the gradients for each layer. Exploding gradients occur due to significantly large gradients compared to the weights. Therefore, the trust ratio will adapt to produce a small value to lower the LR, reducing the chance of divergence. The corresponding effect happens for vanishing gradients.

\subsection{LAMB}
The LAMB algorithm is an instantiation of the layerwise adaptive strategy with the normalization modification performed on the Adam optimizer. In LAMB, there are two normalizations. The first normalization occurs when $m_t^{(i)}$ is normalized with $v_t^{(i)}$, providing adaptivity for each weight. Furthermore, the second normalization occurs layerwise when computing the trust ratio. Despite having two normalizations, the authors of LAMB \cite{you2019lamb} provided convergence guarantees that proves LAMB's convergence. Algorithm \ref{algo:lambc} shows the pseudocode for LAMB.

\section{Methodology}

\begin{algorithm}[t]
\caption{\colorbox{BurntOrange}{LAMB} and \colorbox{YellowGreen}{LAMBC} algorithms}
\label{algo:lambc}
\begin{algorithmic}[1]
\State \textbf{Given:} $w_1\in \mathbb{R}^d$, learning rate policy $\{\eta_t\}_{t=1}^T$, $0<\beta_1,\beta_2<1$, $\phi$, $\epsilon>0$, $h$-layer neural network model $\mathcal{M}$, clipping parameters $c=\{$\colorbox{YellowGreen}{True}, \colorbox{BurntOrange}{False}$\}$, \colorbox{YellowGreen}{$\mu$}
\State \textbf{Initialize:} $m_0=0$, $v_0=0$
\For {$t=1$ \textbf{to} T}
    \State Draw b samples $S_t$ from training set.
    \For {$i=1$ \textbf{to} $h$}
        \State $g_t^{(i)}=\frac{1}{|S_t|}\sum_{s_t\in S_t}\nabla L(w_t^{(i)}, s_t)$
        \State $m_t^{(i)} = \beta_1m_{t-1}^{(i)}+(1-\beta_1)g_t^{(i)}$
        \State $v_t^{(i)} = \beta_2v_{t-1}^{(i)}+(1-\beta_2)(g_t^{(i)})^2$
        \State $m_t^{(i)}={m_t^{(i)}}/{(1-\beta_1)}$
        \State $v_t^{(i)}={v_t^{(i)}}/{(1-\beta_2)}$
    \EndFor
    \State Compute ratio $r_t=\frac{m_t}{\sqrt{v_t}+\epsilon}$
    \State Compute trust ratio $\gamma_t=\frac{\phi(||w_t^{(i)}||)}{||r_t^{(i)}||}$
    \If {$c=$ \colorbox{YellowGreen}{True}}
        \State \colorbox{YellowGreen}{$\gamma_t \leftarrow \texttt{Clip}(\gamma_t, \mu)$}
    \EndIf
    \State $w_{t+1}^{(i)}=w_t^{(i)}-\eta_t\gamma_t^{(i)}(r_t^{(i)}+\lambda w_t^{(i)})$
\EndFor
\end{algorithmic} 
\end{algorithm}

Layerwise adaptive methods, such as LAMB, are observed to  suffer from unstable and extreme trust ratios which degrades performance. This happens when the weight norm becomes too large compared to the gradient norm, leading to divergence during training. To solve this problem, we apply a clipping operation to the trust ratio (ratio between the L2-norm of weights and the L2-norm of the per layer gradients) by constraining it to be within a range of values between a predefined set of upper and lower bounds. Specifically, at any time $t$,

\begin{equation}\label{eq:clip}
\dfrac{||w^{(i)}||}{||\nabla L(w^{(i)})||}=
\begin{cases} 
      \mu & \text{if}\hspace{3mm} \dfrac{||w^{(i)}||}{||\nabla L(w^{(i)})||} < \mu \\
      \tau & \text{if}\hspace{3mm} \dfrac{||w^{(i)}||}{||\nabla L(w^{(i)})||} > \tau \\
      \dfrac{||w^{(i)}||}{||\nabla L(w^{(i)})||} & \text{if}\hspace{3mm} \mu < \dfrac{||w^{(i)}||}{||\nabla L(w^{(i)})||} < \tau
\end{cases}
\end{equation}

where $\mu\in \mathbb{R}^+$ and $\tau\in \mathbb{R}^+$ are the lower and upper bounds for all layers, while $w^{(i)}$ and $\nabla L(w^{(i)})$ are the weights and gradients for layer $i$ respectively. In our implementation, we set the lower bound $\tau=0$. Clipping the trust ratio prevents the weight update from exploding to huge values. As such, the degraded performance or training divergence caused by extreme and unstable trust ratios can be alleviated.

The upper bound $\mu$ and lower bound $\tau$ in the clipping operation in Eq. \ref{eq:clip} are hyperparameters that are manually tuned and are set as constant for all layers. To improve the flexibility of training, one may consider using adaptive methods to decide the upper and lower bound for different layers during training. In our preliminary experiments, we tried an adaptive method from \cite{luo2019adaptive} to investigate this possibility. However, our empirical results show that such adaptive method performs not as well as manually defined bounds, although with improvements over no clipping. We postulate that the dynamics of the upper and lower bound functions in \cite{luo2019adaptive} does not fit the evolution of the trust ratio values during training, resulting in conflicting scenarios whereby clipping is performed on the trust ratio when it should not.

\section{Experiments}
In this section, we will introduce the experiments we have done to validate the performance of the trust ratio clipping. Our experiments aim to answer the following questions:
\begin{enumerate}
    \item Can trust ratio clipping help with the task generalization and test performance?
    
    \item If trust ratio clipping works, what is the best or recommended trust ratio value that we should adopt?
    
    \item Does the trust ratio clipping work in a more complex image classification task such as on ImageNet \cite{deng2009imagenet}?
\end{enumerate}
                                                  
Following the three aforementioned questions, we divide the experiment section into three parts, each with individual experiments that address the respective research questions in detail.

\subsection{Image classification on CIFAR10}
In this section, we aim to find out whether applying trust ratio clipping results in test performance improvement compared against without clipping. We test our hypothesis on the CIFAR10 dataset \cite{krizhevsky2009learning} that contains 60000 32x32 colour images in 10 classes. Due to limited computation resources, we choose ResNet-18 \cite{he2016deep} as our neural network model backbone.  We set the learning rate to be 1e-2, number of epochs to be 80 and compared the performance with and without trust ratio clipping on various batch sizes, ranging from 1000 to 3000. If trust ratio clipping is enabled, we clip the trust ratio to be less than 1.

\begin{figure}[th]
\centering
\includegraphics[width=\linewidth]{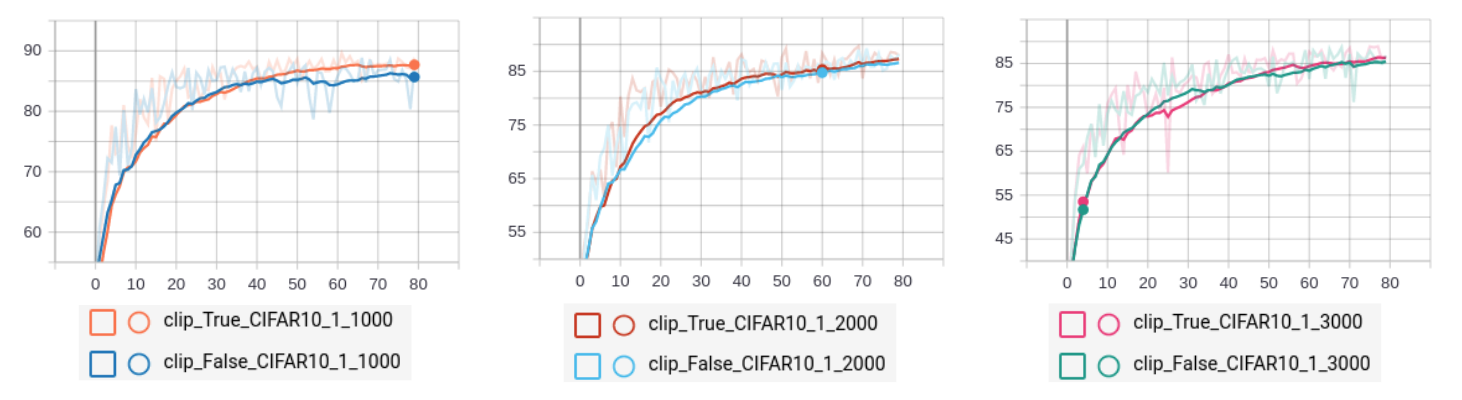}
\caption[HMM]{Image classification task on CIFAR10 dataset \cite{krizhevsky2009learning}. We compare the scenarios with and without the trust ratio clipping. X-axis is the number of epochs and y-axis is the prediction accuracy in percentage. We conduct experiments on different batch sizes: 1000 (left), 2000 (middle) and 3000 (right). }
\label{fig:1}
\end{figure}

\begin{table}[h!]
\centering
\begin{tabular}{ |c|c|c|c| } 
\hline
Batch Size & 1000 & 2000 & 3000 \\
\hline

Test Accuracy (clip) &  \textbf{87.71} & \textbf{87.3} & \textbf{86.29}\\
Test Accuracy (no clip) &  85.68 & 86.61 & 85.41\\
\hline 
\end{tabular}

\caption{Quantitative results for the test performance across different batch sizes. }
\label{table:results}
\end{table}

Figure \ref{fig:1} and Table \ref{table:results} shows the evaluation results on CIFAR 10 dataset. All the three tasks with different batch sizes clearly indicate an improvement brought by trust ratio clipping on the final testing performance. With batch size 1K, trust ratio clipping has the highest improvement of about 2\% for the testing accuracy and about 0.7\% improvement for the other batch sizes. Therefore, we conclude that trust ratio clipping can improve on the task generalization and test performance.

\subsection{Selecting the suitable trust ratio clipping bound}

Observing the success of trust ratio clipping in the previous experiments, we are curious about what is the best or recommended value to clip the trust ratio. We conduct another set of experiments on the CIFAR10 dataset \cite{krizhevsky2009learning} with different upper bound values of trust ratio clipping. We test the trust ratio values on four different scales: 1, 3, 5, 10. 

\begin{figure}[th]
\centering
\includegraphics[width=\linewidth]{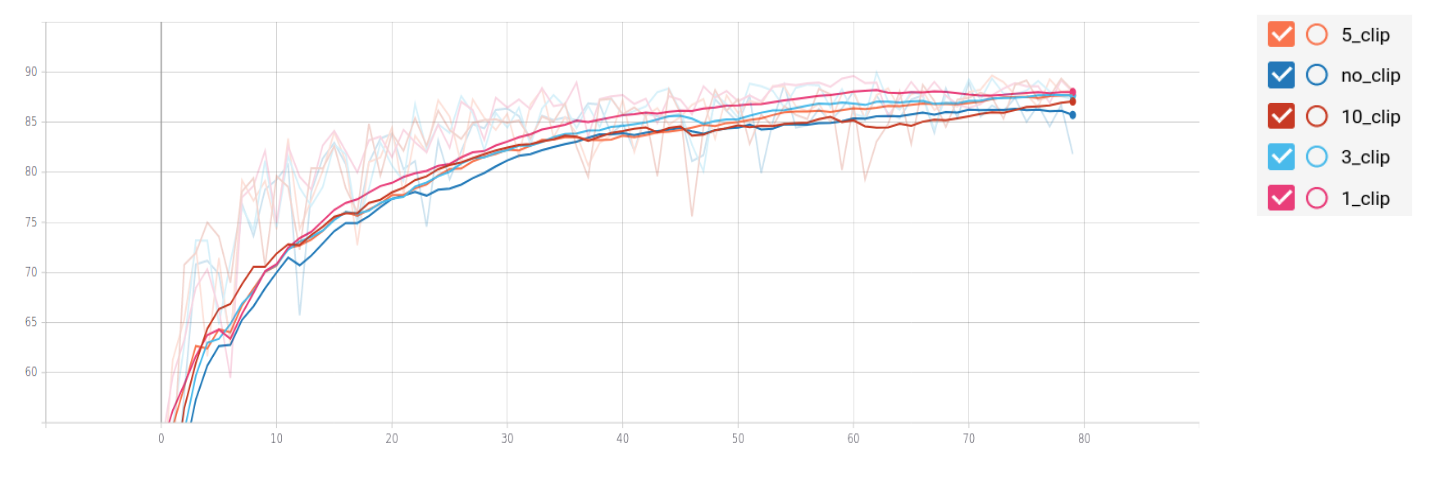}
\caption[HMM]{Image classification task on CIFAR10 dataset \cite{krizhevsky2009learning} with different trust ratio values. X-axis is the number of epochs and y-axis is the prediction accuracy in percentage. }
\label{fig:2}
\end{figure}

Figure \ref{fig:2} shows the evaluation results. Interestingly, we discover that all the conditions with trust ratio clipping outperform the no-clipping setup, while the final testing performance is inversely proportional to the maximum clipping value. From the figure, clipping with max value 1 is the best, 10 is the worst, and 3, 5 sits in between. Since trust ratio reflects the ratio between the magnitudes of the neural network weights and gradients, a possible explanation will be that drastic gradient updates on a stabilized weight parameter may jeopardize the generalization performance. This is indeed true, if the scale of the weight value attempts to stabilize, drastic changes (with trust ratio > 1) on the weight parameter take higher risks to downgrade the generalization performance. This observation is also consistent in the first experiments in figure \ref{fig:1}, where models with clipping only starts to surpass models without clipping at the later stage after the scale of the weight stabilized.  

An insight discovered here is to dynamically adjust the maximum trust ratio clipping value. At the beginning of the training, larger trust ratios should be allowed but it should be avoided after the weight scaling stabilizes, e.g. the converging phase, to prevent the risks of downgrading the generation performance.

\subsection{Image classification on ImageNet}
In this section, we aim to find out whether trust ratio clipping works in more complex image classification tasks such as ImageNet \cite{deng2009imagenet}. ImageNet consists of 14 million real images with a total of 1000 classes. The original dataset occupies about 500G disk space, which exceeds beyond the capability of our computational resources and we have to conduct our experiments on the down-sampled ImageNet dataset on the scale of 64x64x3 per images. The batch size is also limited to the size of 400. 

\begin{figure}[th]
\centering
\includegraphics[width=\linewidth]{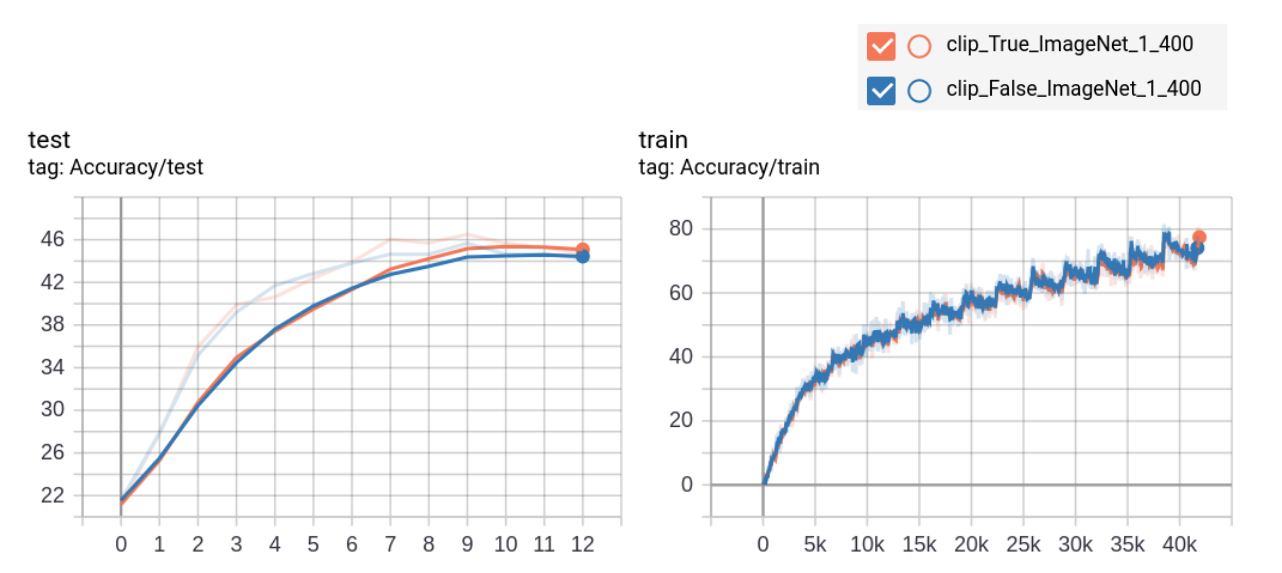}
\caption[HMM]{Image classification task on the ImageNet dataset. X-axis is the number of epochs and y-axis is the prediction accuracy in percentage. }
\label{fig:3}
\end{figure}

The final result is shown in Figure \ref{fig:3}. From the figure, even with the more complex image classification task on ImageNet, our proposed trust ratio clipping still helps with the generalization performance and outperforms the model without trust ratio clipping. Although, our model seems to over-fit the training data with a much higher accuracy in the training dataset, our objective is not to achieve absolute performance in test dataset but to show the effectiveness of the trust ratio clipping over the model without trust ratio clipping.

\section{Future Work}
As observed from the empirical results, it is crucial to define a good clipping bound for the task. Ideally, the selection of the clipping bound should be tuned as accurately as possible. This is to ensure that each weight update can be significant, while the magnitude of the weight update should also just be large enough for controllable and optimal updates. This requires going beyond manual specification of clipping bound values and exploring adaptive methods for the clipping bounds. In our preliminary experiments, we attempted the technique from \cite{luo2019adaptive} and provided an analysis on the possible subpar performance compared against manual clipping. Following this line of thought, we suggest other possible methods for trust ratio adaptivity. Inspired by \cite{ede2020adaptive}, a possible approach is to consider maintaining a standard deviation from the trust ratio and setting the clipping bound $n$ standard deviations away from the mean.

One of the main limitations of our approach is that the same clipping bound value is applied for all layers in the neural network. However, it has been observed in \cite{you2017lars} that the trust ratio values can vary significantly among different layers in the network. Therefore, the direction towards applying an adaptive trust ratio clipping should also take this into consideration and adopt a layerwise trust ratio clipping approach too.

Through our experiments, we have verified LAMBC on image classification tasks with ImageNet and CIFAR-10 datasets. However, we were unable to investigate its effectiveness on large batch sizes ($>$ 8K on CIFAR-10 and $>$ 1K on ImageNet) due to a lack of computational resources. It would be interesting to analyze the effects of clipping on both small and large batch training using LAMBC. Furthermore, we must also test and verify the algorithm's effectiveness on a wider range of tasks, such as language modeling and neural machine translation, etc. It is important that trust ratio clipping must not degrade LAMBC's generalization ability.

\section{Conclusion}
Large batch training is critical to accelerating training of large deep neural networks. The existing approach for large batch training, the LAMB optimizer, features adaptive layerwise learning rates based on computing the trust ratio. Trust ratios explicitly compare the L2-norm of layer weights over the L2-norm of layer gradients, and uses this difference as an adaptive feedback to adjust the overall layerwise learning rate.

However, the trust ratio introduced by LAMB is still vulnerable to extreme gradient values due to the increasing norm of weights of layers within neural networks. The unstable and extreme trust ratio can lead to degrading performance of trained model. To solve this problem, we present, a new variant of LAMB, called LAMBC, that clips the trust ratio corresponding to the predefined clipping bound value. Clipping constrains the trust ratios within a reasonable range of values, which prevents the gradient update from exploding to huge values, while improving the final performance of the trained model by encouraging a reasonable rate of weight update. 

We evaluated LAMBC on image classification tasks using different datasets, including CIFAR-10 and ImageNet. LAMBC achieves a better performance than LAMB for all of the experiments, with better generalization ability and higher test accuracies. LAMBC also works effectively across small and large batch sizes, as well as across different clipping bound values. Although all the investigated clipping bound values improves the performance compared against no clipping, it was observed that the selection of clipping bound value is still paramount to the success of trust ratio clipping. Therefore, training LAMBC with a suitable adaptive trust ratio clipping approach is an immediate future work to look into.

\section*{Acknowledgement}
We would like to thank Yang You for his valuable input regarding potential adaptive trust ratio clipping methods. We also want to thank the National University of Singapore for computational resource support. We would like to acknowledge that this work is done for the course CS6285: Bridging Systems and Deep Learning.

\bibliographystyle{ieeetr}
\bibliography{references}

\end{document}